\documentclass[format=acmsmall]{acmart}

\AtBeginDocument{%
  }

\renewcommand\footnotetextcopyrightpermission[1]{} 
\usepackage{enumitem}
\usepackage{graphicx}
\usepackage[framemethod=TikZ]{mdframed}
\usepackage{xcolor}
\usepackage{comment}

\setlist[itemize]{leftmargin=1em}

\begin{document}

\title{The Indispensable Role of User Simulation in the Pursuit of AGI}

\author{Krisztian Balog}
\affiliation{%
  \institution{University of Stavanger}
  \country{Norway}
}
\email{krisztian.balog@uis.no}

\author{ChengXiang Zhai}
\affiliation{%
  \institution{University of Illinois at Urbana-Champaign}
  \country{USA}
}
\email{czhai@illinois.edu}

\begin{abstract}
Progress toward Artificial General Intelligence (AGI) faces significant bottlenecks, particularly in rigorously evaluating complex interactive systems and acquiring the vast interaction data needed for training adaptive agents. This paper posits that user simulation---creating computational agents that mimic human interaction with AI systems---is not merely a useful tool, but is a critical catalyst required to overcome these bottlenecks and accelerate AGI development. We argue that realistic simulators provide the necessary environments for scalable evaluation, data generation for interactive learning, and fostering the adaptive capabilities central to AGI. Therefore, research into user simulation technology and intelligent task agents are deeply synergistic and must advance hand-in-hand. This article elaborates on the critical role of user simulation for AGI, explores the interdisciplinary nature of building realistic simulators, identifies key challenges including those posed by large language models, and proposes a future research agenda.
\end{abstract}




\maketitle

\section{Introduction}

The rapid advancements in generative artificial intelligence (AI), particularly large language models (LLMs), have fueled significant excitement about the potential for achieving Artificial General Intelligence (AGI)---AI exhibiting human-level cognitive abilities across a wide range of tasks. We observe impressive performance on specific benchmarks and a steady growth in model capacity. Current AGI research  focuses heavily on scaling these foundation models and enhancing specific agent capabilities, such as complex reasoning and coding. However, despite this progress, even the most advanced AI systems remain far from possessing common sense reasoning, planning, and robust generalization capabilities, which are hallmarks of human intelligence.

A critical, yet often underestimated, bottleneck hindering faster progress toward AGI is the heavy reliance on human interaction data for training and evaluation. This process is inherently slow, expensive, and difficult to scale. User simulation---the use of computational models to mimic human behavior during interactions with AI systems---offers a powerful avenue to address this bottleneck. By generating synthetic interaction data and providing automated, scalable, and reproducible evaluation environments, user simulation can significantly accelerate the iterative cycle of AI development and testing, thereby speeding up progress toward AGI. 

However, research dedicated to user simulation has not yet garnered the attention it deserves, given its potential impact on the AGI quest. In this article, we argue that user simulation is not merely a useful tool but a \emph{critical component} for achieving AGI. We posit that the development of highly capable AI agents and the creation of realistic user simulations must proceed hand-in-hand. These two lines of research are synergistic, advancing toward AGI from complementary directions: one focusing on the agent's capabilities, the other on the complexity and realism of the interaction environment it operates within.

Creating realistic user simulations is an inherently interdisciplinary challenge. It requires integrating insights not only from machine learning and natural language processing but also crucially from psychology, cognitive science, and human-computer interaction to accurately model the complexities of human decision-making, preferences, biases, and interaction patterns \citep{Balog:2024:FnTIR}. While recent breakthroughs in LLMs provide powerful new tools for building more sophisticated simulators~\citep{Thomas:2024:SIGIR, Argyle:2023:PA, Park:2023:UIST}, leveraging these tools effectively necessitates a deep, interdisciplinary understanding of human behavior.

\section{What is User Simulation?}

User simulation involves creating computational agents designed to mimic how real humans might interact with an AI system. These agents are built using algorithms, rules, or models informed by our understanding of user behavior, knowledge, preferences, and cognitive processes. Crucially, they can often be parameterized to represent a diverse range of user characteristics, for example, novices vs. experts or users with different goals and interaction styles.

User simulators serve two critical roles when integrated with interactive AI systems: 
(1) they enable \emph{evaluation} via repeatable, reproducible, and low-cost experiments, saving invaluable user time, and (2) they can generate large-scale synthetic interaction data for \emph{training}, especially when real user data is scarce, sensitive, or difficult to obtain.

The scope of user simulation is broad, ranging from relatively simple models predicting single user actions (like clicks or ratings) to sophisticated agents modeling complex, goal-oriented behavior across multiple tasks or sessions (such as writing code).

At a high level, there are two main families of simulation techniques. \emph{Model-based} approaches rely on explicit representations of behavior, such as predefined rules derived from expert knowledge or interpretable probabilistic models capturing uncertainties. These often allow parameters (set heuristically or learned from data) to be tuned to represent different user types. In contrast, \emph{data-driven} approaches leverage machine learning, often deep neural networks, to learn interaction patterns directly from large datasets of observed user behavior. While these models can achieve high predictive fidelity, they operate as ``black boxes,'' sacrificing interpretability regarding why the simulator behaves a certain way. 

Effective user simulators are characterized by several key properties, including validity, interpretability, cognitive plausibility, variation, and adaptability~\cite{Balog:2024:FnTIR}. 
Although optimizing a simulator across all these dimensions is desirable, real-world development usually involves navigating trade-offs tailored to the application. High fidelity might be paramount for data augmentation, for example, while interpretability is often key for evaluation purposes.
Reflecting these trade-offs, hybrid approaches are also common, combining model-based techniques with machine-learned components to balance different properties.

Crucially, simulation does not need to be perfect to be useful.  
In fact, creating a ``perfect'' user simulator, i.e., one that flawlessly replicates human behavior across all possible tasks and contexts, is likely an AI-complete problem, on par with achieving AGI.

\section{User Simulation to Accelerate the Path to AGI}

The development of realistic user simulators is, in many respects, fundamentally aligned with the broader pursuit of AGI---creating intelligent agents with human-like capabilities.
This alignment is not merely conceptual; it is reflected in the shared technological foundations employed by both fields throughout AI history. From rule-based expert systems prevalent in the 1980s and 90s, through the adoption of probabilistic models in the 2000s, the subsequent rise of machine-learned models, and the recent wave of Transformer architectures and LLMs, advancements in core AI technologies have consistently been leveraged to build both more capable task agents \emph{and} more realistic user simulators. 
Consequently, technical challenges in building sophisticated simulators often mirror those in developing intelligent task agents, suggesting deep and synergistic connections between user simulation and AGI research.

Beyond this foundational alignment, user simulation directly addresses critical bottlenecks slowing progress in AI development. It enables scalable, reproducible, and low-cost \emph{evaluation}, significantly accelerating development cycles compared to relying solely on time-consuming, non-reproducible, and expensive human testing. Furthermore, simulation can be used to generate synthetic interaction data, essential for \emph{training} agents (e.g., via reinforcement learning), particularly when real data is scarce, sensitive, or unavailable at the required scale. 
Indeed, the concept of simulation is already implicitly embedded in modern LLM training paradigms: in Reinforcement Learning from Human/AI Feedback, the learned reward model, trained on human/AI-generated preference labels to act as a proxy for human judgment, essentially functions as a non-interpretable user simulator capturing preferences to guide the agent's learning. 
An autonomous intelligent agent may leverage its own user simulation agent to self-generate synthetic data for optimization of its interactions with real users, thus implementing a functional analogue of Theory of Mind~\citep{Premack:1978:BBS}. 

Beyond evaluation and data generation, user simulation is indispensable for developing agents capable of effective \emph{human-AI collaboration}. Realizing the full potential of such partnerships requires AI agents to do more than merely exhibit superhuman task performance; they must account for the inherent variability in their human partners' behavior, including diverse problem-solving approaches, individual preferences, and suboptimal actions.
The importance of this adaptation is underscored by recent work in chess, a domain that has long served as a Petri dish for AI research: human players paired with AI agents tailored to their skill level outperform those partnered with more powerful AI agents that are not adjusted for skill-compatibility~\citep{Hamade:2024:ICLR}.

Achieving such effective, synergistic collaboration necessitates that the AI agent understands, predicts, and adapts to its human partner.
This adaptation fundamentally requires the agent to leverage sophisticated models of the user's knowledge, intentions, and decision-making processes---essentially demanding an embedded or tightly integrated user simulation capability. 
An intelligent agent assisting a user must respond appropriately to user actions and inferred needs based on such a model. This implies a close, reciprocal relationship: the task agent uses simulation for feedback to optimize its interaction policy, while the simulator must potentially adapt to changes in the task agent or environment over time. Therefore, the interdependence between research on intelligent task agents and user simulation is inherent and likely to persist until AGI is achieved.

\section{Advancing User Simulation in the Age of LLMs: Challenges and Roadmap}

The emergence of large language models represents a significant technological leap, potentially accelerating the integration and synergy between intelligent task agents and user simulators discussed earlier. As these models demonstrate increasing capabilities, they may well serve as a foundational building block for both types of agents. Indeed, LLMs have already fueled extensive adoption as both task agents and simulation tools across diverse domains and applications (for comprehensive surveys, see~\citep{Gao:2024:HSSC, Wang:2024:FCS}). However, while this presents a considerable opportunity, harnessing the power of LLMs specifically for \emph{realistic} user simulation requires confronting several significant challenges.

\subsubsection*{\textbf{Challenge: Achieving Realistic and Controllable Behavior}}
While LLMs can produce fluent interactions, their generated responses can be unpredictable, sometimes unsafe, exhibit unrealistic or incoherent patterns of behavior, and may also lack the natural variation observed in real human interactions~\citep{Balog:2024:FnTIR}. 
Furthermore, LLMs often possess more knowledge than average humans and generate overly ``perfect'' responses. This ``superuser'' effect, while beneficial and, indeed, intended for a task agent, results in unrealistic simulations that fail to capture typical human limitations, knowledge gaps, biases, or error patterns. While prompting techniques can guide LLM behavior, ensuring strict adherence to instructions remains a challenge~\citep{Balog:2024:FnTIR}. Therefore, a key direction for future research is to develop more robust methods for controlling and calibrating LLM behavior within simulation contexts. This includes techniques to reliably constrain outputs, define specific personas with realistic limitations (e.g., cognitive capabilities), inject natural variation and error patterns, and calibrate knowledge levels to match target user profiles, moving beyond the limitations of current prompting strategies. 

\subsubsection*{\textbf{Challenge: Bridging the Cognitive Gap}}
Beyond surface behavior, a more fundamental challenge lies in bridging the cognitive gap between current LLM capabilities and human cognition.
While LLMs might be aware of concepts like patience or satisfaction, they lack the training data to model the human dynamics of such behaviors. 
Similarly, LLMs lack a deep understanding of core human cognitive processes, such as decision-making, memory recall, and attention span, and may fail to accurately simulate mechanisms of decision-making under uncertainty, the limitations of working memory, shifts in attention, or the influence of cognitive biases. These shortcomings hinder their ability to generate realistic simulations of human users. 
To overcome this, LLMs must be extended with components that capture a wider range of human cognitive abilities. The human brain is known to consist of two distinct systems: System 1, which is intuitive, fast, but not reliable, and System 2, which is logical, deliberate, but slower~\cite{Daniel:2013:Book}. While current LLMs appear to be able to simulate System 1 well, they lack the deliberate reasoning and planning capabilities characteristic of System 2.  
To address this cognitive gap, we need to explore hybrid architectures that integrate explicit cognitive models---capturing aspects like memory update mechanisms, attention allocation strategies, or decision heuristics identified in cognitive science---with the generative power of LLMs. 
Neurosymbolic approaches, which combine neural learning's flexibility with symbolic reasoning's structure and interpretability, represent a particularly promising direction for embedding ``System 2 capabilities''  into user simulators~\cite{Garcez:2023:AIReview}.

\subsubsection*{\textbf{Challenge: Fostering Interdisciplinary Research and Community}}

Building simulators that reflect authentic human behavior requires more than just sophisticated machine learning. 
We need insights from \emph{psychology}, \emph{cognitive science}, and \emph{human-computer interaction} for empirically grounded models of interaction patterns, cognitive processes (like attention, memory limitations, and biases), and the diversity of user populations. Effective modeling of user goals, evolving knowledge during interaction, and reasoning capabilities necessitates leveraging techniques from \emph{information science} and \emph{knowledge representation}. Furthermore, connections with \emph{intelligent agents} and \emph{multi-agent systems} are crucial for simulating complex social interactions and potentially embedding simulation capabilities within adaptive task agents. 

Therefore, a critical future direction is actively building bridges between these disciplines. This requires creating and supporting venues to foster interdisciplinary collaborations, and developing shared experimental platforms and evaluation resources. Fostering a vibrant, truly interdisciplinary research community dedicated to user simulation is paramount for making substantial progress.

\section{Conclusion}
The journey toward AGI demands more than advances in agent capabilities alone; it requires robust methods for evaluation, training, and ensuring adaptive interaction. This paper has made the case that user simulation is a critical, indispensable technology for meeting these demands, highlighting its synergistic interdependence with core AGI agent research. 
Therefore, we advocate for a renewed focus and strategic investment in user simulation research as crucial for making progress toward truly intelligent systems.

\bibliographystyle{ACM-Reference-Format}
\bibliography{references}


\begin{thebibliography}{10}


\ifx \showCODEN    \undefined \def \showCODEN     #1{\unskip}     \fi
\ifx \showDOI      \undefined \def \showDOI       #1{#1}\fi
\ifx \showISBNx    \undefined \def \showISBNx     #1{\unskip}     \fi
\ifx \showISBNxiii \undefined \def \showISBNxiii  #1{\unskip}     \fi
\ifx \showISSN     \undefined \def \showISSN      #1{\unskip}     \fi
\ifx \showLCCN     \undefined \def \showLCCN      #1{\unskip}     \fi
\ifx \shownote     \undefined \def \shownote      #1{#1}          \fi
\ifx \showarticletitle \undefined \def \showarticletitle #1{#1}   \fi
\ifx \showURL      \undefined \def \showURL       {\relax}        \fi
\providecommand\bibfield[2]{#2}
\providecommand\bibinfo[2]{#2}
\providecommand\natexlab[1]{#1}
\providecommand\showeprint[2][]{arXiv:#2}

\bibitem[Argyle et~al\mbox{.}(2023)]%
        {Argyle:2023:PA}
\bibfield{author}{\bibinfo{person}{Lisa~P. Argyle}, \bibinfo{person}{Ethan~C.
  Busby}, \bibinfo{person}{Nancy Fulda}, \bibinfo{person}{Joshua~R. Gubler},
  \bibinfo{person}{Christopher Rytting}, {and} \bibinfo{person}{David
  Wingate}.} \bibinfo{year}{2023}\natexlab{}.
\newblock \showarticletitle{Out of One, Many: Using Language Models to Simulate
  Human Samples}.
\newblock \bibinfo{journal}{\emph{Political Analysis}} \bibinfo{volume}{31},
  \bibinfo{number}{3} (\bibinfo{year}{2023}), \bibinfo{pages}{337--351}.
\newblock


\bibitem[Balog and Zhai(2024)]%
        {Balog:2024:FnTIR}
\bibfield{author}{\bibinfo{person}{Krisztian Balog} {and}
  \bibinfo{person}{ChengXiang Zhai}.} \bibinfo{year}{2024}\natexlab{}.
\newblock \showarticletitle{User Simulation for Evaluating Information Access
  Systems}.
\newblock \bibinfo{journal}{\emph{Foundations and Trends in Information
  Retrieval}} \bibinfo{volume}{18}, \bibinfo{number}{1-2}
  (\bibinfo{year}{2024}), \bibinfo{pages}{1--261}.
\newblock


\bibitem[Daniel(2013)]%
        {Daniel:2013:Book}
\bibfield{author}{\bibinfo{person}{Kahneman Daniel}.}
  \bibinfo{year}{2013}\natexlab{}.
\newblock \bibinfo{booktitle}{\emph{Thinking, fast and slow}}.
\newblock \bibinfo{publisher}{New York : Farrar, Straus and Giroux}.
\newblock


\bibitem[Gao et~al\mbox{.}(2024)]%
        {Gao:2024:HSSC}
\bibfield{author}{\bibinfo{person}{Chen Gao}, \bibinfo{person}{Xiaochong Lan},
  \bibinfo{person}{Nian Li}, \bibinfo{person}{Yuan Yuan},
  \bibinfo{person}{Jingtao Ding}, \bibinfo{person}{Zhilun Zhou},
  \bibinfo{person}{Fengli Xu}, {and} \bibinfo{person}{Yong Li}.}
  \bibinfo{year}{2024}\natexlab{}.
\newblock \showarticletitle{Large Language Models Empowered Agent-based
  Modeling and Simulation: A Survey and Perspectives}.
\newblock \bibinfo{journal}{\emph{Humanities and Social Sciences
  Communications}}  \bibinfo{volume}{11}, Article \bibinfo{articleno}{1259}
  (\bibinfo{year}{2024}).
\newblock


\bibitem[Garcez and Lamb(2023)]%
        {Garcez:2023:AIReview}
\bibfield{author}{\bibinfo{person}{Artur~d’Avila Garcez} {and}
  \bibinfo{person}{Luis~C Lamb}.} \bibinfo{year}{2023}\natexlab{}.
\newblock \showarticletitle{Neurosymbolic AI: The 3rd wave}.
\newblock \bibinfo{journal}{\emph{Artificial Intelligence Review}}
  \bibinfo{volume}{56}, \bibinfo{number}{11} (\bibinfo{year}{2023}),
  \bibinfo{pages}{12387--12406}.
\newblock


\bibitem[Hamade et~al\mbox{.}(2024)]%
        {Hamade:2024:ICLR}
\bibfield{author}{\bibinfo{person}{Karim Hamade}, \bibinfo{person}{Reid
  McIlroy-Young}, \bibinfo{person}{Siddhartha Sen}, \bibinfo{person}{Jon
  Kleinberg}, {and} \bibinfo{person}{Ashton Anderson}.}
  \bibinfo{year}{2024}\natexlab{}.
\newblock \showarticletitle{Designing Skill-Compatible {AI}: Methodologies and
  Frameworks in Chess}. In \bibinfo{booktitle}{\emph{The Twelfth International
  Conference on Learning Representations}} \emph{(\bibinfo{series}{ICLR '24})}.
\newblock


\bibitem[Park et~al\mbox{.}(2023)]%
        {Park:2023:UIST}
\bibfield{author}{\bibinfo{person}{Joon~Sung Park}, \bibinfo{person}{Joseph
  O'Brien}, \bibinfo{person}{Carrie~Jun Cai}, \bibinfo{person}{Meredith~Ringel
  Morris}, \bibinfo{person}{Percy Liang}, {and} \bibinfo{person}{Michael~S.
  Bernstein}.} \bibinfo{year}{2023}\natexlab{}.
\newblock \showarticletitle{Generative Agents: Interactive Simulacra of Human
  Behavior}. In \bibinfo{booktitle}{\emph{Proceedings of the 36th Annual ACM
  Symposium on User Interface Software and Technology}}
  \emph{(\bibinfo{series}{UIST '23})}. Article \bibinfo{articleno}{2}.
\newblock


\bibitem[Premack and Woodruff(1978)]%
        {Premack:1978:BBS}
\bibfield{author}{\bibinfo{person}{David Premack} {and} \bibinfo{person}{Guy
  Woodruff}.} \bibinfo{year}{1978}\natexlab{}.
\newblock \showarticletitle{Does the chimpanzee have a theory of mind?}
\newblock \bibinfo{journal}{\emph{Behavioral and Brain Sciences}}
  \bibinfo{volume}{1}, \bibinfo{number}{4} (\bibinfo{year}{1978}),
  \bibinfo{pages}{515--526}.
\newblock


\bibitem[Thomas et~al\mbox{.}(2024)]%
        {Thomas:2024:SIGIR}
\bibfield{author}{\bibinfo{person}{Paul Thomas}, \bibinfo{person}{Seth
  Spielman}, \bibinfo{person}{Nick Craswell}, {and} \bibinfo{person}{Bhaskar
  Mitra}.} \bibinfo{year}{2024}\natexlab{}.
\newblock \showarticletitle{Large Language Models can Accurately Predict
  Searcher Preferences}. In \bibinfo{booktitle}{\emph{Proceedings of the 47th
  International ACM SIGIR Conference on Research and Development in Information
  Retrieval}} \emph{(\bibinfo{series}{SIGIR '24})}.
  \bibinfo{pages}{1930--1940}.
\newblock


\bibitem[Wang et~al\mbox{.}(2024)]%
        {Wang:2024:FCS}
\bibfield{author}{\bibinfo{person}{Lei Wang}, \bibinfo{person}{Chen Ma},
  \bibinfo{person}{Xueyang Feng}, \bibinfo{person}{Zeyu Zhang},
  \bibinfo{person}{Hao Yang}, \bibinfo{person}{Jingsen Zhang},
  \bibinfo{person}{Zhiyuan Chen}, \bibinfo{person}{Jiakai Tang},
  \bibinfo{person}{Xu Chen}, \bibinfo{person}{Yankai Lin},
  \bibinfo{person}{Wayne~Xin Zhao}, \bibinfo{person}{Zhewei Wei}, {and}
  \bibinfo{person}{Jirong Wen}.} \bibinfo{year}{2024}\natexlab{}.
\newblock \showarticletitle{A Survey on Large Language Model based Autonomous
  Agents}.
\newblock \bibinfo{journal}{\emph{Frontiers of Computer Science}}
  \bibinfo{volume}{18} (\bibinfo{year}{2024}).
\newblock


\end{thebibliography}

\end{document}